# Layer-Wise Evolution of Representations in Fine-Tuned Transformers: Insights from Sparse AutoEncoders

Suneel Nadipalli

**Abstract**

Fine-tuning pre-trained transformers is a powerful technique for enhancing the performance of base models on specific tasks. From early applications in models like BERT to fine-tuning Large Language Models (LLMs), this approach has been instrumental in adapting general-purpose architectures for specialized downstream tasks. Understanding the fine-tuning process is crucial for uncovering how transformers adapt to specific objectives, retain general representations, and acquire task-specific features. This paper explores the underlying mechanisms of fine-tuning, specifically in the BERT transformer, by analyzing activation similarity, training Sparse AutoEncoders (SAEs), and visualizing token-level activations across different layers. Based on experiments conducted across multiple datasets and BERT layers, we observe a steady progression in how features adapt to the task at hand: early layers primarily retain general representations, middle layers act as a transition between general and task-specific features, and later layers fully specialize in task adaptation. These findings provide key insights into the inner workings of fine-tuning and its impact on representation learning within transformer architectures.

## 1. Introduction

Since their introduction in 2017 [1], transformers have driven significant advancements in Natural Language Processing (NLP). Tasks such as question answering, machine translation, and summarization have seen remarkable progress, largely enabled by the capabilities of transformers. One of the most impactful advancements is fine-tuning, which allows pre-trained models to be adapted for specific downstream tasks, achieving performance levels comparable to training from scratch—without requiring the same computational resources [2]. Both transformers and fine-tuning have found widespread applications, particularly in the field of Large Language Models (LLMs). While transformers form the backbone of powerful LLMs, fine-tuning further refines their functionality for specialized tasks.

Despite the widespread use of fine-tuning in both LLMs and broader NLP applications, the underlying mechanisms of the process remain underexplored. Questions such as how fine-tuning modifies learned representations, which layers play key roles in adaptation, and why certain fine-tuning tasks fail on specific models are still open research questions [3] [4]. Gaining these insights could provide a deeper understanding of how pre-trained transformers modify and retain representations during fine-tuning, offering valuable information about the broader learned representations within transformers. Additionally, it could help identify which transformers are more compatible with specific tasks, improving task-to-model alignment within the context of fine-tuning.

Moreover, in high-stakes applications such as medical and financial NLP, understanding how fine-tuning alters model representations is crucial—not only for improving performance but also for ensuring accountability and reliability [5] [6] [7]. A better grasp of the fine-tuning process could explain model behavior, helping practitioners assess why certain pre-trained transformers perform the way they do on specialized tasks.

Beyond tracking feature progression across layers, another important aspect of fine-tuning is how different types of linguistic knowledge evolve during the process. Prior work suggests that low-level syntactic features (such as word order, part-of-speech dependencies) tend to be captured in earlier layers, while high-level semantic understanding (such as meaning shifts, task-specific reasoning) emerges in later layers. This raises the question: Are certain types of representations more resistant to fine-tuning than others? By analyzing which features remain stable versus which undergo significant transformation, we can gain deeper insights into the trade-offs between general linguistic retention and task-specific adaptation.

This study aims to track the progression of learned features during fine-tuning by comparing the pre-trained and fine-tuned representations of BERT across its layers. Inspired by [8], we leverage Sparse AutoEncoders (SAEs) to extract and analyze the evolving feature space, enforcing mono-semanticity in learned representations to make complex activation patterns more interpretable [9]. By visualizing token-level activations and comparing them across different datasets, we aim to uncover how fine-tuning reshapes representations at different depths within the model.

Through our experiments, we observe that early layers in BERT retain general representations, with the 3rd layer capturing broad semantic structures that remain relatively unchanged across fine-tuning, suggesting that these layers encode fundamental linguistic features transferable across tasks. The middle layers act as a transition zone, where by the 6th layer, BERT begins adapting to the fine-tuning task while still preserving some general features, indicating a gradual shift from general-purpose representations to task-specific ones. And finally, later layers, such as the 12th layer, fully specialize in the fine-tuning task, largely discarding general linguistic features in favor of representations that are highly relevant to the specific task at hand.

## 2. Related Work

Recent work in understanding transformer representations has focused on how features evolve across layers and how fine-tuning alters these learned representations. Studies on mechanistic interpretability have shown that transformer layers exhibit a structured progression, with lower layers capturing general syntactic properties, middle layers encoding a mix of syntax and semantics, and deeper layers specializing in task-specific features [10] [11] [12]. Prior works such as [13] have further demonstrated that BERT's fine-tuned layers encode task-relevant transformations while retaining core semantic properties. However, most analyses have focused on static, pre-trained models rather than tracking how these representations shift during fine-tuning. Prior research using techniques like Canonical Correlation Analysis (CCA) and Singular Value Canonical Correlation Analysis (SVCCA) [14] has provided insights into representation similarity across models, revealing that early layers remain relatively stable while later layers undergo significant change. Yet, these methods primarily compare the similarity of activations across models and training stages, exploring how much of the overall structure is preserved. However, they do not explicitly look at the individual concepts learned across these training stages, identifying which of them are retained, modified or forgotten.

The effects of fine-tuning on learned representations have also been explored within the context of catastrophic forgetting, with studies examining how transformers overwrite general-purpose knowledge when adapting to new tasks [15]. While some research has shown that careful fine-tuning strategies—such as parameter-efficient tuning—can preserve general representations [16] [17], there remains limited work analyzing this at a feature level. Instead of treating fine-tuning as a black-box optimization process, understanding how features transition layer-by-layer could provide deeper insights into why certain tasks benefit more from fine-tuning than others and why some models struggle to adapt without degrading performance on prior tasks.

More recent advances in feature analysis have leveraged dictionary learning and Sparse AutoEncoders (SAEs) to extract monosemantic features from transformer activations. For instance, the work done in [18] has demonstrated that these SAEs can decompose transformer activations into interpretable components by disentangling complex representations. It indicates that SAEs can be used to isolate individual features, making the interpretation of specific concepts across different layers that much easier.

Similarly, open-source SAEs have been developed and trained on various layers of transformer models. This initiative aims to help researchers understand the inner workings of language models by providing tools to analyze how features evolve throughout the model and interact to form more complex features [19].

Further research in this area has shown that SAEs can decompose transformer activations into interpretable components, enforcing mono semanticity and improving the disentanglement of mixed representations [9]. These findings highlight how SAEs can isolate individual monosemantic features, making it easier to interpret specific concepts across different layers of transformers. By applying SAEs to large language models, this research has demonstrated that learned representations can be broken down into distinct, human-interpretable units. This work [20] builds on prior research in dictionary learning but extends it by focusing on scaling interpretability techniques to modern transformer architectures, proving SAEs to be a powerful tool for driving interpretability research.

Additionally, studies on transformer circuits have demonstrated that model activations can be decomposed into structured computational units [21]. This approach has helped clarify how specific neurons and attention heads contribute to model behavior. Complementary work has also shown that transformer feed-forward layers serve as key-value memories that store structured features relevant to downstream tasks [22]. These findings suggest that feature retention in transformers follows a structured mechanism, further motivating the need for fine-grained analysis of representation shifts during fine-tuning.

While previous studies have provided valuable insights into representation learning for transformers, there is still a lack of comprehensive, layer-wise analysis of the effect of fine-tuning on feature retention and adaptation. This work aims to build on these previous works and bridge this gap by using SAEs to analyze concepts across pre-trained and fine-tuned BERT models, providing a clearer picture of features evolving across layers and tasks.

## 3. Methodology

### 3.1 Data and Models

The base model used in these experiments was *bert-base-uncased*, a 12-layer transformer pre-trained on a large corpus of unstructured text. To analyze how fine-tuning affects learned representations, the model was fine-tuned on three distinct datasets, each representing a different type of NLP task. The goal was to observe how BERT adapts to these tasks and each task's effect on the fine-tuning process.

Each dataset underwent basic text preprocessing, including converting all text to lowercase, removing punctuation marks, and filtering out special characters. Target labels were mapped to numerical values where necessary to ensure compatibility with the training process. Finally, each dataset was truncated to approximately 25,000 rows and split into 80% training, 10% validation, and 10% testing.

The datasets used in this study include the IMDb dataset, which consists of movie reviews labeled with sentiments (positive or negative) for binary sentiment analysis. The Spotify dataset contains user reviews of the Spotify app, with ratings from 1 to 5, making it suitable for multi-class classification. And lastly, the News dataset comprises news headlines categorized into the following topics: wellness, food, politics, entertainment, and sports. The performances of BERT fine-tuning on each of these datasets can be seen in Table 2 from **Appendix A.**

And finally, to track feature progression during fine-tuning in a computationally feasible manner, analysis was limited to three key layers of BERT's 12-layer architecture. These layers were chosen as they represent pivotal points in the model's transformation process: Layer 3 (early layer), Layer 6 (middle layer) and Layer 12 (late layer). This rationale is supported by works like [12] and [23], which establish that the layers in BERT do indeed exhibit a pipeline-like behaviour, with more and more complex linguistics tasks being captured by deeper and deeper layers.

### 3.2 Experiment Setup

#### 3.2.1 FINE-TUNING

The *bert-base-uncased* model was fine-tuned separately on each of the three datasets to analyze how its internal representations evolve during task adaptation, for 3 distinct types of tasks. Training was conducted using the Adam optimizer and a learning rate of $2*10^{-5}$ over 10 epochs. The pre-training and post-fine-tuning performances have been recorded in Table 2 from **Appendix A**.

#### 3.2.2 EXTRACTING ACTIVATION VALUES

To establish a baseline hypothesis on how features evolve across different layers during fine-tuning, activation values were extracted and compared between pre-trained and fine-tuned models. For each dataset, the token matrix of every sample was passed through all 12 layers of the *bert-base-uncased* model. This process was also repeated for three additional BERT variants of varying model sizes: *bert-medium* (8 layers), *bert-small* (4 layers), and *bert-tiny* (2 layers). The activation matrix (output of each layer) was extracted using the model's *hidden_states* variable. To make comparison and visualization easier, the activation values for each sample were averaged, producing a single representative activation vector for each dataset-layer-size combination.

For each of these combinations, the cosine similarity was computed between the corresponding activation vectors of the pre-trained and fine-tuned versions. Previous studies, such as [24] and [23], used cosine similarity to analyze changes in model representations, and found it very effective for tracking feature retention and transformation across layers. These similarity scores were then plotted across multiple configurations, providing an easy, visual and intuitive way to analyze how fine-tuning affects feature retention and transformation at different depths of the model.

This step helped establish a baseline hypothesis of how features evolve across a BERT transformer in general. Performing the experiment and visualizing the results for various configurations of BERT helped reinforce this hypothesis.

#### 3.2.3 TRAINING SAEs

The next step was to test the established hypothesis in a more detailed and systematic manner. Building off prior work on Sparse AutoEncoders (SAEs) in interpretability [9, 13], we trained SAEs to analyze how learned features evolve across different layers of the model.

Following a similar process as in Section 3.2.2, activation matrices were extracted for the 3rd, 6th, and 12th layers of the pre-trained and fine-tuned models across each dataset. These activation matrices served as the training data for the SAEs. The architecture of each SAE consisted of:
- An **encoder** with an input dimension of 768 (matching the hidden state size of BERT) and an output dimension of 1024 (expanding feature space to capture richer representations).
- A **decoder** with an input and output dimension of 1024, reconstructing the transformed activation space.

The loss function used for training combined Mean Squared Error (MSE) and a sparsity loss term, defined as:

$$L_{sparsity} = \lambda \sum_i |h_i|$$

Where $\lambda$ is a sparsity regularization parameter set to $10^{-3}$ in this experiment. This encourages the activation vectors to be sparse, meaning that each neuron activates for a more selective and interpretable subset of features. The final combined loss function is given by:

$$L_{loss} = L_{MSE} + L_{sparsity}$$

The SAEs were trained using the Adam optimizer with a learning rate of $2*10^{-5}$ over 10 epochs. They demonstrated strong performance, consistently reaching a loss on the order of $10^{-4}$.

By enforcing sparsity, the SAEs forced activations toward monosemanticity, making them less complicated and more interpretable as discrete, human-understandable concepts.

3.2.4 ANALYSING TOKEN- LEVEL ACTIVATIONS

With the SAEs trained, the next step was to run experiments on the extracted features and interpret them as human-understandable concepts. To achieve this, *token-level activations* were computed for selected test sentences, allowing for a deeper analysis of what each learned feature represents.

The first step involved extracting the learned features from the SAEs, specifically focusing on the top n most variable features—those with the highest activation variation across the dataset. Given an activation matrix where each row corresponds to a feature index and each cell represents the activation value for the ith feature and jth sample, the variance was computed along the sample dimension. Features were then ranked in descending order by variance, with the most variable features selected for further analysis. The reasoning behind this selection is that high-variance features are more likely to represent distinct, meaningful concepts, rather than noise or redundant information. For this experiment, the top 3 features were chosen for interpretation. For each selected feature index, token-level activations were plotted for a diverse set of test sentences. The process involved:

1. Tokenizing the sentence and passing it through the corresponding layer of the *bert-base-uncased* model.

2. Extracting the **hidden state activations** from that layer.

3. Passing these activations through the corresponding trained SAE to obtain the activation values for the **selected feature index.**

4. Recording the **activation values for each token** in the sentence.

By plotting these token-wise activations, the experiment allowed for a rigorous analysis of which tokens activated a given feature the most. Testing across a range of sentence types—from general-purpose sentences to those containing domain-specific vocabulary—provided insight into the specific linguistic or semantic concept that a particular feature index captures.

This process was conducted comparatively for both pre-trained and fine-tuned models, analyzing how the concepts learned at each layer-feature index combination evolved through fine-tuning. Comparing these concepts between models helped establish the final hypothesis regarding how feature representations flow and transform during the fine-tuning process.

## 4. Results

### 4.1 Fine-Tuning Classification Reports

The fine-tuned models were evaluated by calculating their accuracies and their classification reports.

For the IMDb dataset, the model had strong performance for both the positive and negative class.

The **positive** class had a precision of 0.91, recall of 0.96 while the **negative** class had a precision of 0.95 and recall of 0.89. This shows that the fine-tuned BERT model effectively captured sentiment-related features and is well suited for this binary classification task.

The news dataset also had high performances all across the board, doing particularly well for the **sports** and **politics** classes (with near perfect scores of 1.00 precision and recall in **sports** and 1.00 precision and 0.94 recall in **politics**). While certain other classes had slightly lower scores (**entertainment** and 0.71 precision and 0.83 recall), this model proved to be generally robust for the topic classification task.

And finally, the classification report for the Spotify dataset indicates a varied performance across the labels. The model had the highest scores for **5**, whereas the model exhibited significantly poorer performance for intermediary labels **2** and **3** where recall values fell to 0/40 and 0.33 respectively. This indicates the model struggled with possibly more ambiguous labels like **2** and **3**, which could share many sentiments compared to more extreme ratings like **1** and **5.** Detailed classification reports for each dataset can be seen in Figure 6 from **Appendix B**.

### 4.2 Activation Similarities

To establish a base hypothesis of feature progression across fine-tuning, the **cosine similarities** were measured across layers for both pre-trained and fine-tuned models. Fig 1 and Fig 2 depict the layer-wise cosine similarities for the averaged activation values for the bert-base and bert-medium models, respectively, across the three datasets: IMDb, Spotify and News.

In Fig 1 (bert-base), cosine similarity values begin at approximately 1.0 in the initial layers, indicating that the early layers retain a strong overlap between pre-trained and fine-tuned activations. As the layers progress, the similarity gradually decreases, with a notable drop-off in the deeper layers, especially beyond layer 10, where task-specific features start to emerge. Among the datasets, the News dataset exhibits the steepest decline in similarity, reflecting a significant shift in representations due to the additional complexity of the topic classification task. In contrast, the IMDb and Spotify datasets show more gradual decreases, suggesting that fine-tuning on sentiment and multi-class classification tasks require fewer modifications to existing representations in comparison.

These trends remain consistent for the bert-medium model variant, as can be seen in Fig 2, with the earlier layer starting off at extremely high similarity that then gradually decreases with significant drop offs at the later layers. A similar contrastive drop off is noted for the News dataset as well. However, with the change in model depth (12 layers -> 8 layers), the overall similarity values are slightly higher across all layers compared to the base model, suggesting that the smaller depth of the medium model limits the extent of feature transformation. A similar trend is observed for the smaller variants as well, with the overall similarity values getting increasingly higher as the number of layers decreases.

The results help establish our hypothesis that early layers retain general linguistic features, while deeper layers undergo substantial adaptation to accommodate task-specific requirements. The comparative analysis between the two models further underscores how task complexity and model depth influence the degree of representation shifts during fine-tuning.

### 4.3 Learned Concepts from SAEs

The trained SAEs offered a detailed lens into the concepts encoded by the BERT-base-uncased model, both in its pre-trained and fine-tuned forms, across different layers and datasets. These observations revealed a clear progression of feature specialization, aligning with the hypothesis that early layers focus on general representations, middle layers act as transitional zones, and later layers specialize in task-specific features. The differences between the pre-trained and fine-tuned models underscore how fine-tuning modifies internal representations to align with specific task requirements (Table 1.)

In the third layer, the pre-trained model predominantly activates on general linguistic elements such as prepositions, articles, and other structural words, while showing minimal sensitivity to domain-specific or sentiment-related concepts. For example, in the IMDb dataset, this layer exhibited sparse activations for emotionally charged or domain-specific terms. However, after fine-tuning, the layer became more task-specific, with high activations observed for domain-relevant terms like "cast," "plot," and "director." Additionally, the polarity of sentiment—positive or negative—was reflected in the activation values, demonstrating that the layer had begun to encode emotional features. Similarly, for the Spotify dataset, fine-tuning shifted the layer's focus from sparse, general activations to features aligned with sentiment and domain-specific terms like "rock," "happy," and "sad." For the News dataset, this layer, which initially activated sparsely and only for emotional words, began to encode both domain-relevant and general semantic features after fine-tuning.

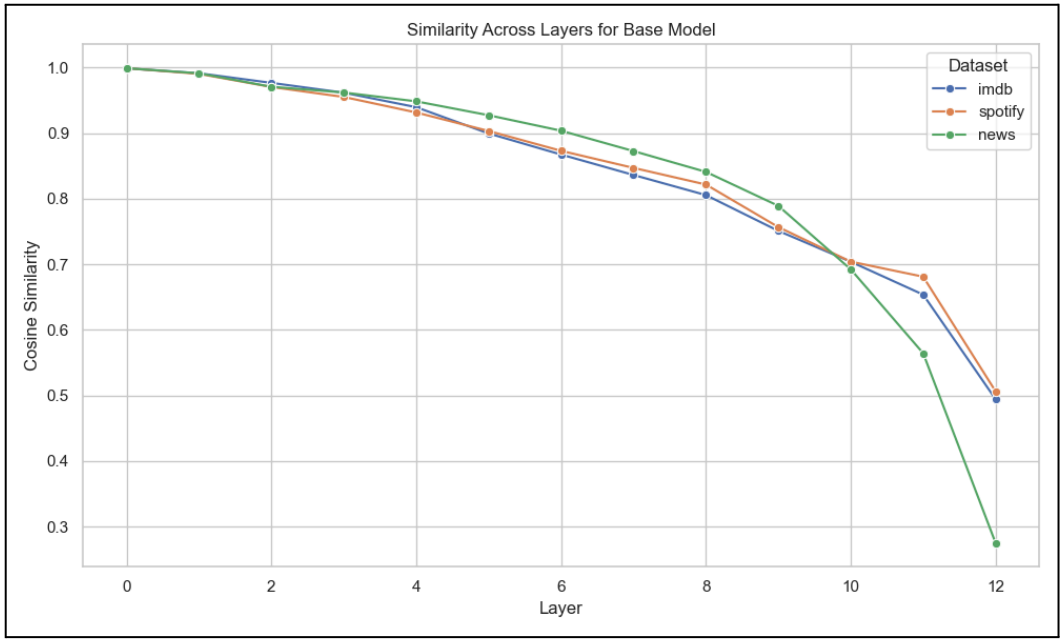

*Fig 1. Summary Graph of Activation Similarities for bert-base Model*

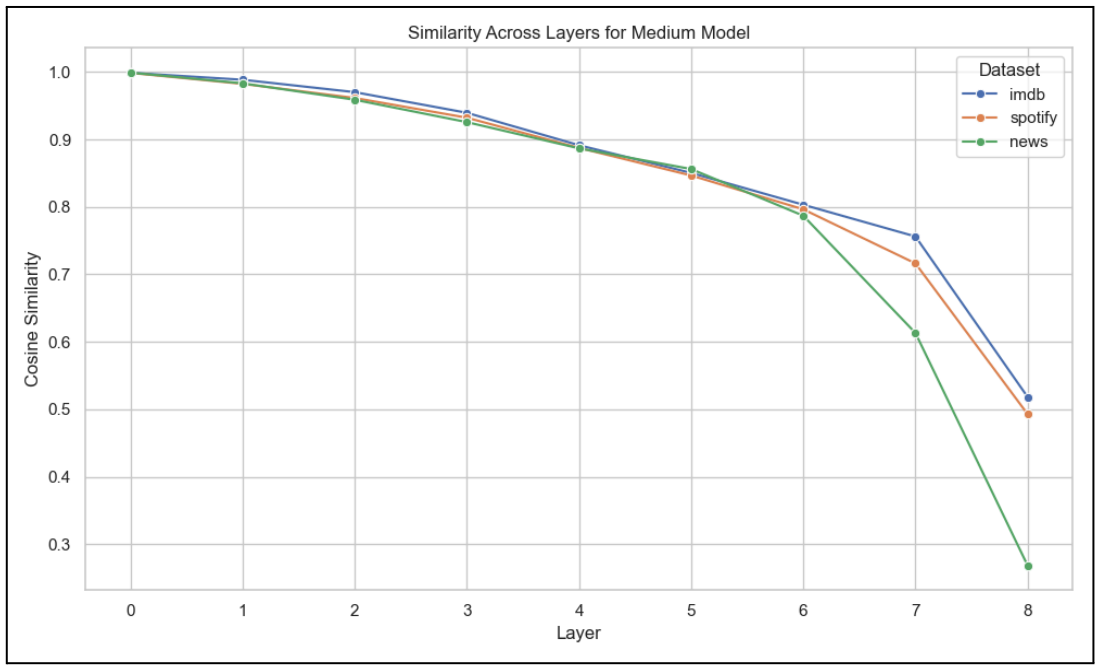

*Fig 2. Summary Graph of Activation Similarities for bert-medium Model*

By the sixth layer, the pre-trained model exhibited a mix of general and high-level semantic features, but its activations often lacked clear, task-specific patterns. For instance, in the IMDb dataset, this layer broadly activated on general and sentiment-related terms, such as "love" and "hate," and showed high sensitivity to structural elements like the final words in sentences. Fine-tuning refined these patterns further, with the model encoding both general and domain-specific words while maintaining consistent activations for punctuation tokens like [CLS] and [SEP]. Interestingly, the model displayed heightened activations for words associated with negative sentiment, signaling a deeper alignment with task requirements. This could also point to a deeper assignment of certain layers to certain sentiments/labels. In the Spotify dataset, the sixth layer evolved to encode a blend of sentiment and domain-specific words, such as "music" and "app," reflecting the interplay between emotional and contextual features. Similarly, for the News dataset, the sixth layer captured domain-specific semantics while reducing focus on sentiment markers, reflecting its role as a transitional layer.

The twelfth layer in the pre-trained model showed relatively uniform activations, with minimal differentiation for task-specific features. However, fine-tuning led to significant transformations in this layer, resulting in highly specialized representations tailored to the respective tasks. In the IMDb dataset, the layer demonstrated a strong ability to distinguish between positive and negative sentiment, with high contrast observed in the activation values for words representing these polarities. Similarly, in the Spotify dataset, the twelfth layer began encoding a range of emotional intensity, clearly differentiating between complex sentiments while maintaining high activations for specific tokens like [SEP]. For the News dataset, this layer, which initially focused on general words with limited task-specific adaptation, began encoding domain-specific features, particularly for the "food" category, while maintaining reduced sensitivity to sentiment.

These observations highlight the systematic progression of learned features across layers in both pre-trained and fine-tuned models. Fine-tuning not only enhanced the model's ability to focus on domain-relevant features but also demonstrated a clear shift from general to task-specific representations as the layers deepened. The findings reinforce the hypothesis that early layers retain general linguistic features, middle layers transition between general and task-specific representations, and later layers specialize entirely in the target task. Figures 3 - 5 plot token-level activations for specific test sentences that highlight additional observations.

### 4.3.1 EARLY LAYERS - LAYER 3

Figures 3A and 3B show the activations for the sentence "The Director of this movie is terrible" on the IMDb dataset. In the pre-trained model, activations are sparse and limited to only a single word, reflecting the minimal task-specific focus of this early layer. By contrast, the fine-tuned model activates for nearly all words in the sentence, with particularly strong activations for domain-specific words like "director" and "movie."

Similarly, Figures 3C and 3D display activations for the sentence "The updates are horrible, the app became worse" on the Spotify dataset. The pre-trained model exhibits sparse activations across just a few words, indicating a lack of task-specific awareness. After fine-tuning, the model demonstrates a more balanced activation profile, still activating for general words but with notable increases for domain-relevant terms like "updates."

Figures 3E and 3F depict the activations for the sentence "10 Of The Best Yoga Poses For Sleep" on the News dataset. Consistent with the trends observed in the other datasets, the pre-trained model produces sparse activations for only a few words, while the fine-tuned model activates broadly across most tokens. Domain-specific words like "poses" and "sleep" show particularly high activation values, further emphasizing the fine-tuned model's focus on task-relevant concepts.

From these observations, it can be concluded that Layer 3 in the pre-trained model primarily encodes general linguistic representations, with minimal semantic focus. Fine-tuning significantly enhances its ability to capture domain-specific and task-relevant features, while retaining some general representations.

### 4.3.2 MIDDLE LAYERS - LAYER 6

Figures 4A and 4B show the activations for the sentence "I love this movie!" on the IMDb dataset. Unlike Layer 3, the pre-trained model at this depth activates consistently across all tokens, particularly on the final word, indicating a stronger representation of general semantics. However, the fine-tuned model displays a shift, with sparse activations concentrated on structural tokens like [CLS] and [SEP]

Table 1. *Summary of the learned features via SAEs for both pre-trained and fine-tuned models*

| Layer | Dataset | Pre-Trained Model | Fine-Tuned Model |
|---|---|---|---|
| **3** | IMDb | Activates on general words (e.g., prepositions, articles); very sparse for domain-specific words. | Activates highly on domain-specific words (e.g., "cast," "plot," "director"); emotional polarity (+ve/-ve) influences activation values. |
|  | Spotify | Sparse activations; does not show task-specific focus. | Activates on sentiment and domain words (e.g., "rock," "happy," "sad"). |
|  | News | Activates sparsely unless emotional words are mentioned. | Activates for domain and general semantic words, with sparse activation for sentiment markers. |
| **6** | IMDb | Activates on general and sentiment words (e.g., "love," "hate"); activates highly on last words. | Encodes both general and domain-specific words; activates consistently on punctuation (e.g., CLS, SEP); higher activation for -ve sentiment words. |
|  | Spotify | Activates broadly for most words without clear patterns. | Encodes combined effects of domain words (e.g., music," app) and sentiment markers. |
|  | News | Represents high-level sentiment context and general semantics. | Activates on domain words and general semantics; reduced focus on sentiment markers. |
| **12** | IMDb | Activates similarly across sentences but shows small variations for +ve/-ve words. | Strongly differentiates between positive and negative sentiment with high contrast in activation values. |
|  | Spotify | Activates highly on emotional words and the [SEP] token | Encodes a range of emotional intensity, clearly distinguishing positive and negative sentiment. |
|  | News | Focuses on general words with minimal task-specific adaptation. | Encodes domain features, specifically the food class |

A similar pattern is observed in Figures 4C and 4D, which display activations for the sentence "I love this app" on the Spotify dataset. While the pre-trained model activates uniformly across most words, the fine-tuned model reduces its activations, focusing heavily on the [SEP] token and other punctuation.

Figures 4E and 4F for the sentence "Talking to Yourself: Crazy or Crazy Helpful?" on the News dataset continues this trend. The pre-trained model maintains activations across most words in the sentence, reflecting its role in encoding general semantics. However, the fine-tuned model activates sparingly, primarily focusing on punctuation and structural tokens.

These results indicate that **Layer 6 in the pre-trained model places a strong focus on encoding general semantic information**, making it an essential transitional layer. In the fine-tuned model, however, this layer shifts its focus, prioritizing **structural tokens** such as punctuation and start/stop markers.

4.3.3 LATE LAYERS - LAYER 12

In the final layer, the focus shifts dramatically toward task-specific representations, as can be seen in Figures 5A and 5B. These figures compare the activations for two sentences from the IMDb dataset: "The plot was unpredictable and the acting was brilliant." (positive sentiment) and "The plot was predictable and the acting was terrible." (negative sentiment). The fine-tuned model shows a clear contrast: while activations are strong for the negative sentiment sentence, the positive sentence exhibits no significant activations. This suggests that Layer 12 in the fine-tuned model has encoded sentiment polarity, prioritizing negative sentiment. In comparison, the pre-trained model activates consistently across most words, showing no sentiment-specific bias.

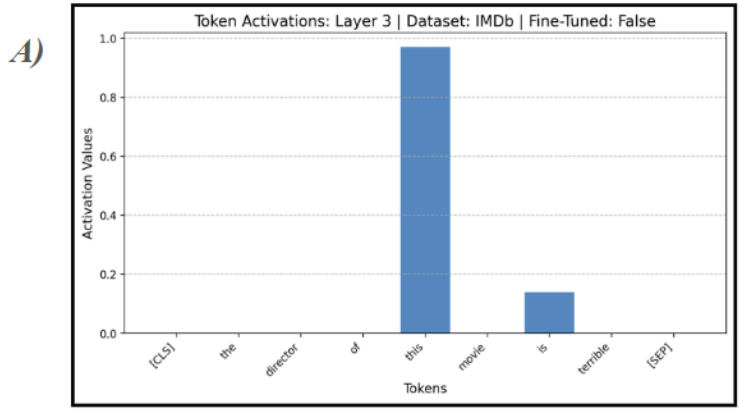
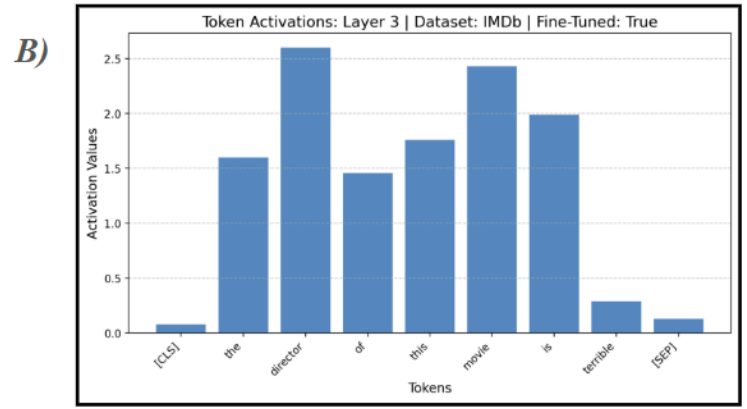
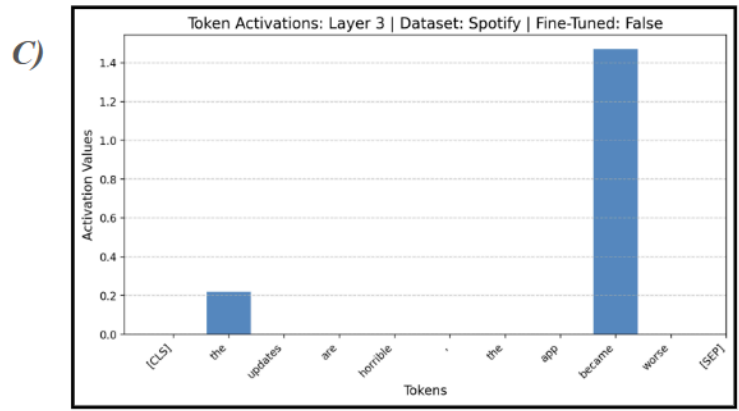
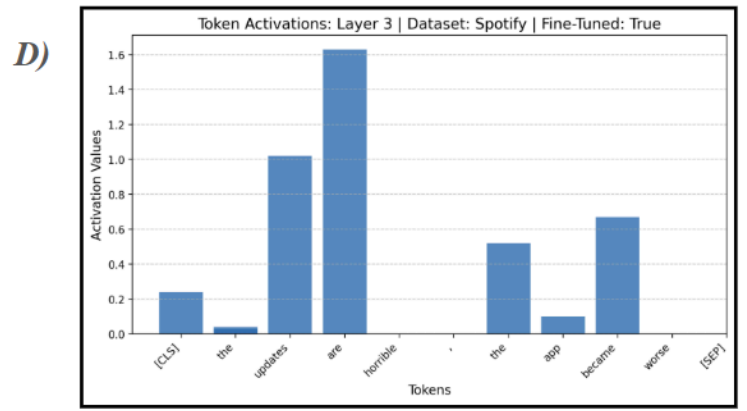
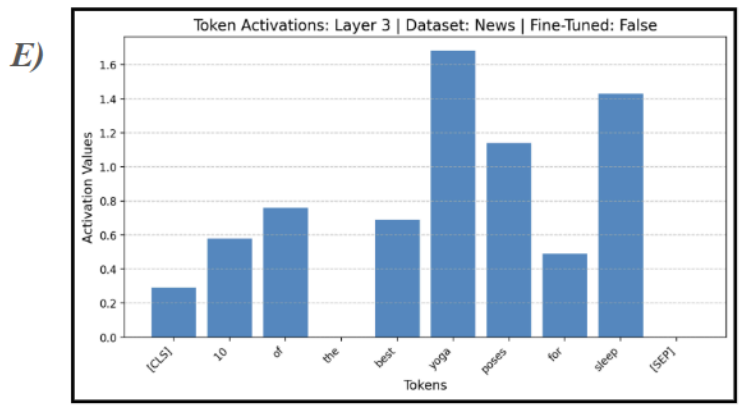
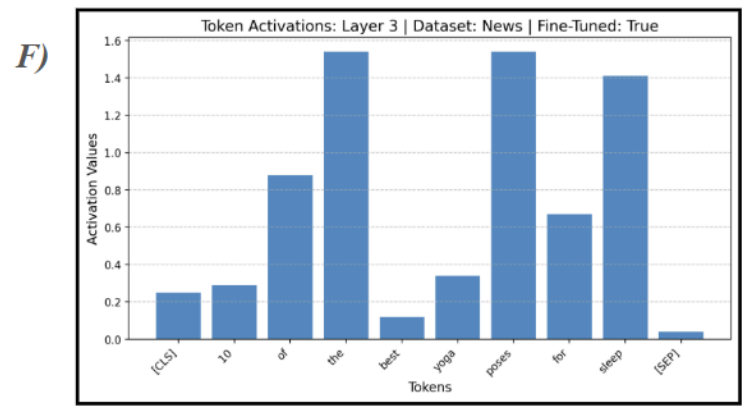

*Fig 3. Activations for different configurations and examples for Layer 3 of BERT*

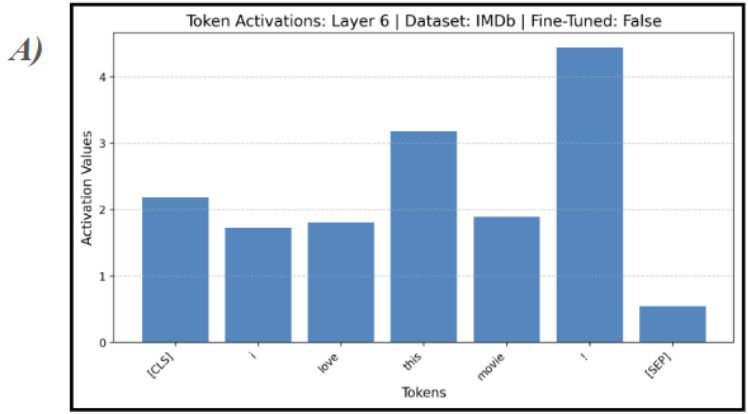 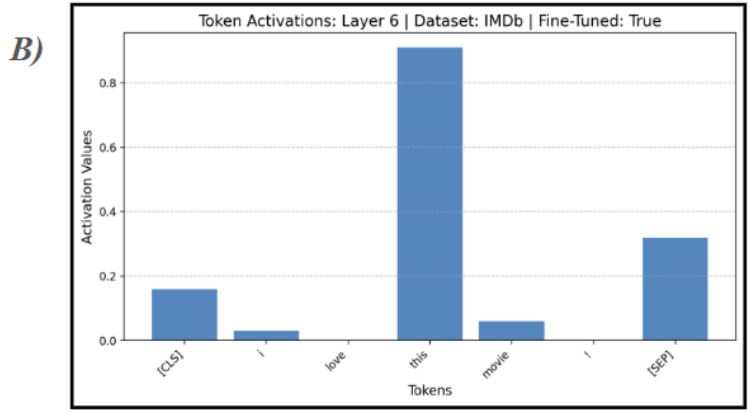 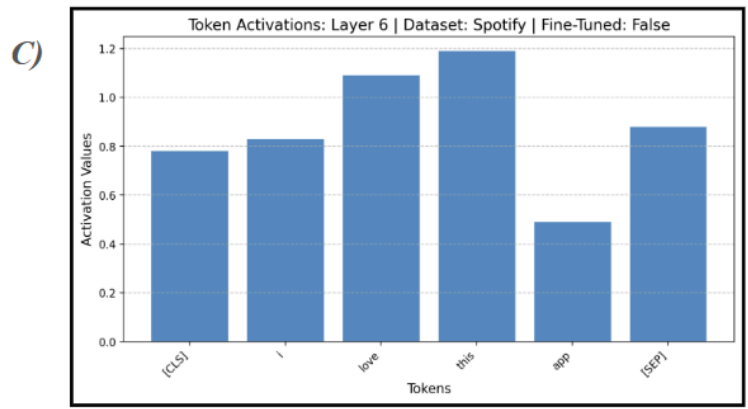 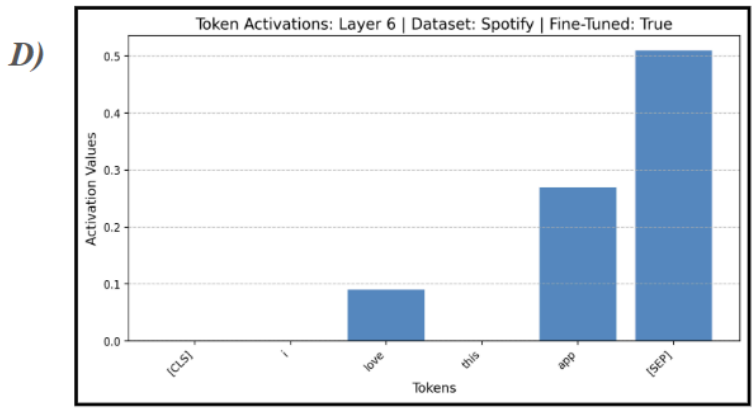 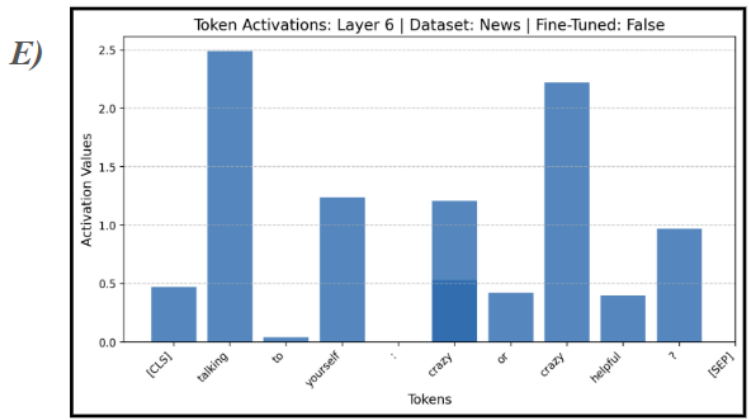 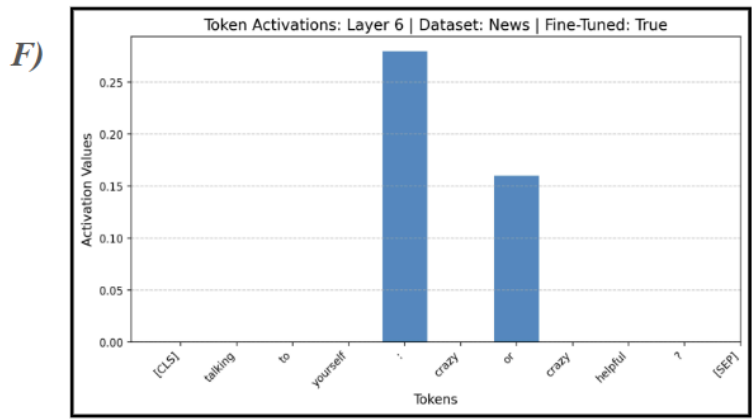

*Fig 4. Activations for different configurations and examples for Layer 6 of BERT*

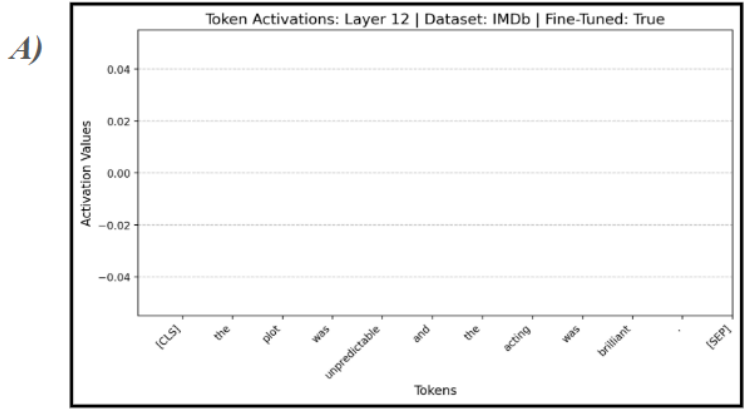
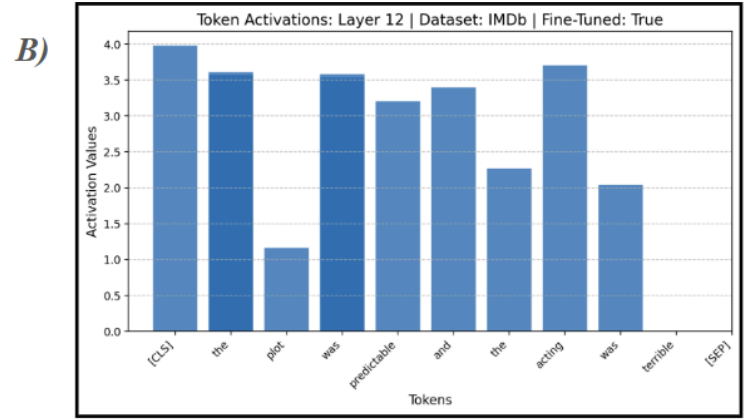
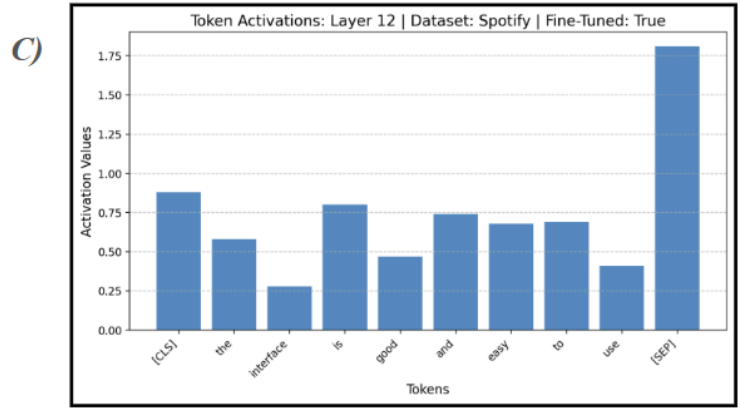
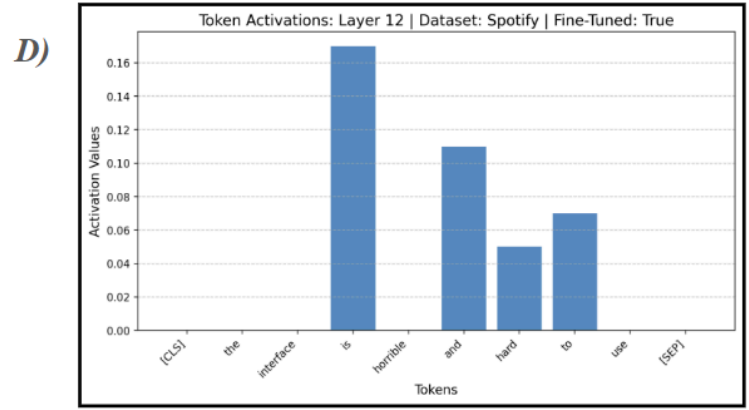
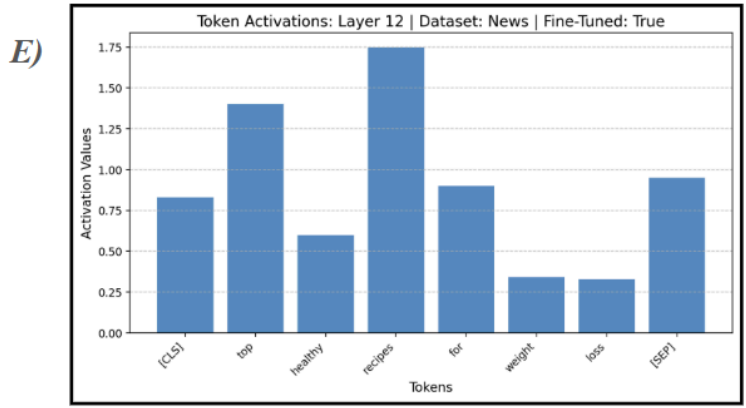
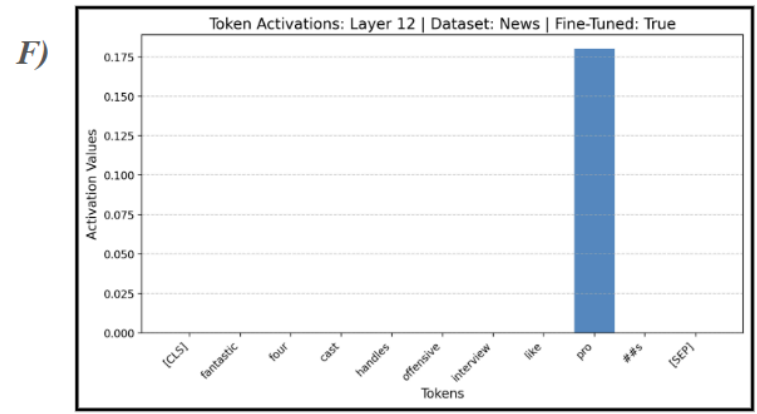

Fig 5. *Activations for different configurations and examples for Layer 12 of BERT*

Figures 5C and 5D illustrate similar findings for the Spotify dataset, with the sentences "The interface is good and easy to use." and "The interface is horrible and hard to use." While the pre-trained model activates broadly across all words, the fine-tuned model displays contrasting activation patterns. Positive sentiment words like "good" and "easy" evoke strong activations in the positive sentence, whereas the negative sentence shows sparse activations across the board, further emphasizing the fine-tuned model's task adaptation.

Finally, Figures 5E and 5F examine two sentences from the News dataset: "Top Healthy Recipes For Weight Loss" and "Fantastic Four Cast Handles Offensive Interview Like Pros." In the pre-trained model, activations are consistent across most words, indicating its general focus. The fine-tuned model, however, shows strong activations for domain-specific features, particularly related to the food category (e.g., "recipes," "healthy"). For the non-food-related sentence, activations are sparse, reflecting the model's specialization in task-relevant classes like "food."

In summary, Layer 12 fully leans into task-specific adaptation. The pre-trained model continues to activate broadly and uniformly, encoding general semantics, while the fine-tuned model demonstrates precise, targeted activations aligned with task requirements.

## 5. Conclusion

This study set out to explore how fine-tuning alters the internal representations of the bert-base-uncased model, with a focus on tracking feature progression across layers and understanding the concepts learned at each stage. Through a combination of cosine similarity analysis, Sparse AutoEncoder (SAE) training, and token-level activation experiments, we uncovered a structured progression in how features are transformed during fine-tuning.

The results demonstrated that **early layers (e.g., Layer 3)** retain general linguistic representations in the pre-trained model but begin to incorporate domain-specific and task-relevant features after fine-tuning. **Middle layers (e.g., Layer 6)** serve as transitional zones, encoding a mix of general semantics and task-specific features in the pre-trained model, but shifting focus to structural markers like punctuation and start/stop tokens in the fine-tuned model. Finally, **late layers (e.g., Layer 12)** show minimal task-specific adaptation in the pre-trained model, while fine-tuning transforms them into highly specialized representations, prioritizing task-relevant features such as sentiment polarity or domain-specific class distinctions.

These findings reinforce the hypothesis that fine-tuning systematically adapts pre-trained representations, moving from general-purpose knowledge in earlier layers to highly specialized features in the later layers. The combination of cosine similarity analysis and SAE-trained feature extraction provided clear insights into the dynamics of this transformation, revealing the nuanced interplay between general retention and task-specific adaptation.

By capturing these patterns, this work contributes to a deeper understanding of the fine-tuning process and provides a framework for analyzing how transformers adapt to diverse downstream tasks. This layer-wise breakdown not only sheds light on the representational shifts in fine-tuned transformers but also sets the stage for more informed and efficient fine-tuning practices in the future.

## 6. Future Work

While this study provides valuable insights into how fine-tuning transforms learned representations in transformers, several avenues remain open for further exploration to deepen our understanding and extend the applicability of these findings.

One promising direction involves expanding the scope of models and tasks. Future experiments could incorporate a wider range of transformer architectures, such as RoBERTa, GPT models, and T5, to evaluate whether the trends observed in this study generalize across different model designs and pre-training objectives. Additionally, testing fine-tuning on a broader variety of tasks, including multimodal tasks, generative tasks, and low-resource settings, could help validate and refine the hypotheses regarding feature evolution and task adaptation. Understanding how these dynamics change in more complex or resource-constrained scenarios would offer critical insights into the versatility of fine-tuning as a technique.

Another key area for exploration is the development of larger and more expressive Sparse AutoEncoders. Increasing the capacity of SAEs could enable more fine-grained monosemanticity, leading to clearer separations between features and enhancing interpretability. Investigating how the size of the autoencoder and the choice of sparsity penalties impact the representations learned would provide a better framework for defining task-relevant features. This could also open doors to optimizing SAEs for specific applications, tailoring their design to the unique demands of different tasks or datasets.

Finally, future work could focus on analyzing cases of poor task adaptation. By applying the same methodologies to tasks where fine-tuning results in suboptimal performance, we could identify differences in feature evolution compared to tasks that align well with the pre-trained transformer representations. Such studies could uncover whether these limitations stem from intrinsic architectural constraints or are dataset-specific challenges, offering potential solutions to improve adaptation for difficult tasks. Understanding these failure modes is critical to building more robust and adaptable transformer models.

By addressing these directions, future research can build on the findings of this study, providing a more comprehensive understanding of fine-tuning dynamics and further advancing the development of interpretable and efficient transformer-based models.

# APPENDIX A - FINE-TUNING RESULTS FROM BERT

**Table 2 -** *Performance Comparison of BERT Model across different datasets*

| Dataset | Task | Pre-Trained | Fine-Tuned |
| --- | --- | --- | --- |
| IMDb | Sent. Analysis | 54% | 93% |
| Spotify | Multi Class. | 12% | 57% |
| News | Topic Class. | 12% | 93% |

# APPENDIX B - CLASSIFICATION REPORTS

**Fig 6 -** *Classification Reports for each dataset as a result of fine-tuning on BERT*

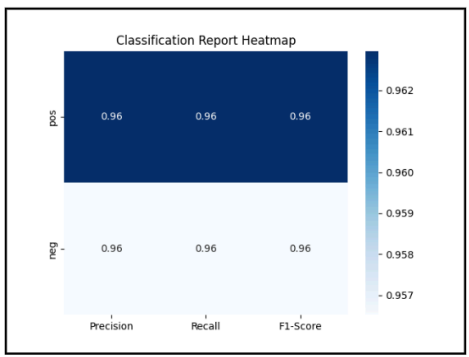

*Fig 6.1 Classification Report for IMDB*

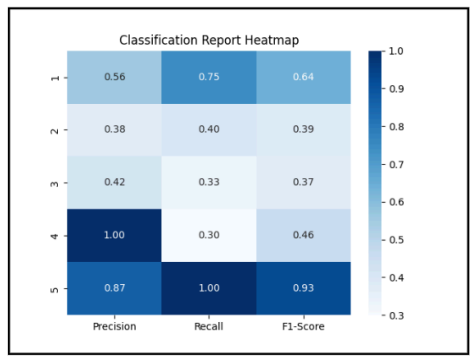

*Fig 6.2 Classification Report for Spotify*

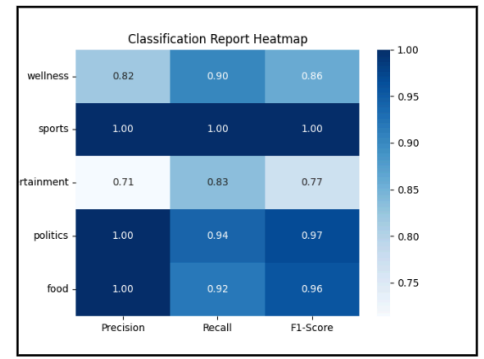

*Fig 6.3 Classification Report for News*